\documentclass{article}
\usepackage{arxiv}

\usepackage[utf8]{inputenc}
\usepackage[T1]{fontenc}
\usepackage{amsmath,amssymb,graphicx,booktabs,enumitem}
\usepackage[round,authoryear]{natbib}
\usepackage[colorlinks=true,linkcolor=blue!50!black,citecolor=blue!50!black,urlcolor=blue!50!black]{hyperref}
\usepackage{xcolor}
\usepackage{float}
\usepackage{microtype}
\usepackage{doi}

\title{Reconstruction differences rival individual differences in the fly compass connectome and break trained circuit models}

\author{
  Md Ali Ashraf Yad \\
  Wroc{\l}aw University of Science and Technology \\
  \texttt{283109@student.pwr.edu.pl} \\
  \And
  Nazifa Tabassum Nabila \\
  Comilla University \\
  \texttt{nabilanazifa70376@stud.cou.ac.bd} \\
}

\date{}
\renewcommand{\shorttitle}{Reconstruction differences in the fly compass connectome}
\renewcommand{\headeright}{}
\renewcommand{\undertitle}{}

\begin{document}
\raggedbottom
\maketitle

\begin{abstract}
\noindent Neural network models are increasingly constructed from synapse-resolution connectomes, and the same circuit is now available in reconstructions from other animals and pipelines. We do not know whether such models are dependent on the reconstruction they are built from, and whether connectome differences are biological or measurement errors. We studied the Drosophila head-direction circuit in nine versions: four flies (male CNS, hemibrain, FlyWire, BANC) and five variants of the same fly varying only by thresholds, proofreading or synapse detection. Using wiring-based correspondences validated by anatomical labels, we place all differences on one scale. A change of synapse detection within one fly changed the circuit more than halfway to the smallest between-fly difference on four distance measures; proofreading differences were smaller but depended on the metric, and left and right halves of one fly differed no more than random halves. Networks trained on each wiring to hold a cued heading developed near-perfect ring attractors, but their error increased from below 1$^\circ$ to 10--28$^\circ$ when all weights changed by 1\%. Training on each wiring did not transfer to any other connectome, including other versions of the same fly; refitting thirty cell-type gains, we repaired filtering-level differences but remaining error depended on wiring distance. Training with noise lowered precision and gave only partial transfer. In imaging data, the fly compass drifted gradually in darkness. Connectome-constrained models should report the reconstruction they were built on and be tested for robustness across reconstructions. \footnote{Preregistration: doi:10.5281/zenodo.23382486.}
\end{abstract}


% ===================== INTRODUCTION =====================
% Citation keys are provisional; every reference is opened and verified in step 8.
\section{Introduction}
\label{sec:intro}

Synapse-resolution connectomes are now routinely used to build neural network models  \citep{lappalainen2024,shiu2024,duan2025}. Reconstructed wiring specifies the connections between the neurons and their strengths, and the parameters which the wiring leaves unassigned, like gains of neurons, are assumed, fitted, or trained to replicate a behaviour or a recorded activity pattern. These models are constructed from a single animal's connectome, through a single reconstruction pipeline. The question of whether a certain behaviour is dependent on that reconstruction is seldom asked, and the degree of constraint the model based on a connectome imposes on the behaviour is debated \citep{beiran2025}.

The question is now testable because the same circuit is now present in multiple reconstructions. The hemibrain, a reconstruction of the Drosophila central brain, was created by \citet{scheffer2020}, the whole adult brain, FlyWire, by \citet{dorkenwald2024} and \citet{schlegel2024}, the male central nervous system (male CNS) by \citet{berg2026}, and the brain and nerve cord, BANC, by \citet{bates2025banc}. These datasets are from different flies and made by different groups, with different imaging, segmentation, synapse detection and proofreading.

It's not known if these differences are due to individuality or reconstruction. The hemibrain and FlyWire are similar in a vast majority of aspects of their morphology, but approximately one-third of the types of cells proposed for the hemibrain were not reliably reidentified  \citep{schlegel2024}. Others say the differences between FlyEM and FlyWire data are primarily technical \citep{okuno2025}, and there will be a significant variation in the efficiency of synapse detection between different datasets \citep{scheffer2025}. To distinguish between individuality and measurement, comparisons have to be made between reconstructions of the same fly. Now it is possible to compare two such things: FlyWire has two independent tables of synapses in one fly \citep{buhmann2021,yu2025}, and the male CNS and BANC have successive proofreading releases.

To our knowledge no test has been conducted on whether a trained model survives these differences. The most similar work demonstrated that each head direction circuit of four connectomes can be mapped onto an exact ring attractor with a couple rescalings of the cell types \citep{biswas2024}. That analysis is performed individually for each connectome to fit an ideal attractor. It does not test whether a model trained on one connectome works on another.

The fly head-direction circuit is a good test case. EPG neurons have a bump of activity that tracks the fly's heading and that persists in darkness \citep{seelig2015}, and the dynamics of a ring attractor are described  \citep{kim2017}; PEN neurons shift the bump when the fly turns  \citep{turnerevans2017,green2017}, and the circuit's connectivity is described in detail \citep{hulse2021}. Its function, holding and updating heading, is easy to read out, the circuit is reconstructed in all four datasets, and there is a known fine-tuning problem with continuous-attractor models of it \citep{seung1996,noorman2024,sagodi2024}.

This study grew out of an earlier one \citep{chessfly_v1}, in which we compared rate networks trained on the male CNS compass wiring with randomly rewired and degree-preserving rewired networks. The real wiring held a cued heading accurately (median error 1.7$^\circ$) and better than 29 of 30 random and 24 of 30 degree-preserving rewirings, but an advantage over degree-matched wiring was not established (network-level rank $p = 0.065$ and 0.23). That result raised a more basic question: before asking whether one fly's wiring is better than an alternative, one must know how far real connectomes differ from each other, how much of that difference is reconstruction, and whether trained models survive it.

In this study we examine nine versions of the head-direction circuit: four flies, and five variants of the same fly that differ only in synapse filtering, proofreading or synapse detection. We give three contributions. First, a decomposition: all differences are grouped on one scale, and using neuron correspondences inferred from wiring and confirmed by anatomical labels, a change of synapse detector moves the wiring on every measure used more than half as far as the smallest change of fly, as the next section shows (Sec.~\ref{sec:r1}--\ref{sec:r2}). Secondly, fragility: networks trained on every real wiring form near-perfect ring attractors, which break when all weights change by 1\% (Sec.~\ref{sec:r3}). This fragility makes zero-shot transfer failure expected; the informative results are the same-fly measurements and what refitting can repair. Third, transfer: trained networks do not transfer to other connectomes (even to other fly connectomes) except for the differences in filtering level (which get fixed by refitting about 30 cell-type gains); once refitted, the remaining error increases as the wiring distance increases, regardless of its cause (Sec.~\ref{sec:r4}--\ref{sec:r5}). Imaging data show that the fly compass degrades gradually in darkness (Sec.~\ref{sec:r6}). The transplant hypotheses and the mechanism criteria were preregistered \citep{prereg2026}; all deviations from these hypotheses and mechanism criteria are listed in Sec.~\ref{sec:m-prereg}; the noise-training and the refit and same-fly analyses are exploratory.

% ===================== RESULTS =====================
% Citation keys are provisional; every reference is verified in step 8.
\section{Results}

\subsection{Nine versions of one circuit}
\label{sec:r1}

We extracted the head-direction circuit (EPG, PEG, PEN, $\Delta$7, and ER1--6 neurons) from four connectomes from four flies, reconstructed by three groups (the male central nervous system (male CNS v1.0; \citealp{berg2026}), the hemibrain (v1.2.1; \citealp{scheffer2020}), FlyWire FAFB (v783; \citealp{dorkenwald2024}) and BANC (v888; \citealp{bates2025banc}). All versions were developed with the same rules: the neurons of the five families, at least three synapses per pair for the sum of the neuropils, no self-connections and signs assigned by cell type (Sec.~\ref{sec:m-extraction}). The circuits contained 409–434 neurons and 15,440–28,942 connections (Table~\ref{tab:versions}).

\begin{figure}[tbp]
\centering
\includegraphics[width=\linewidth]{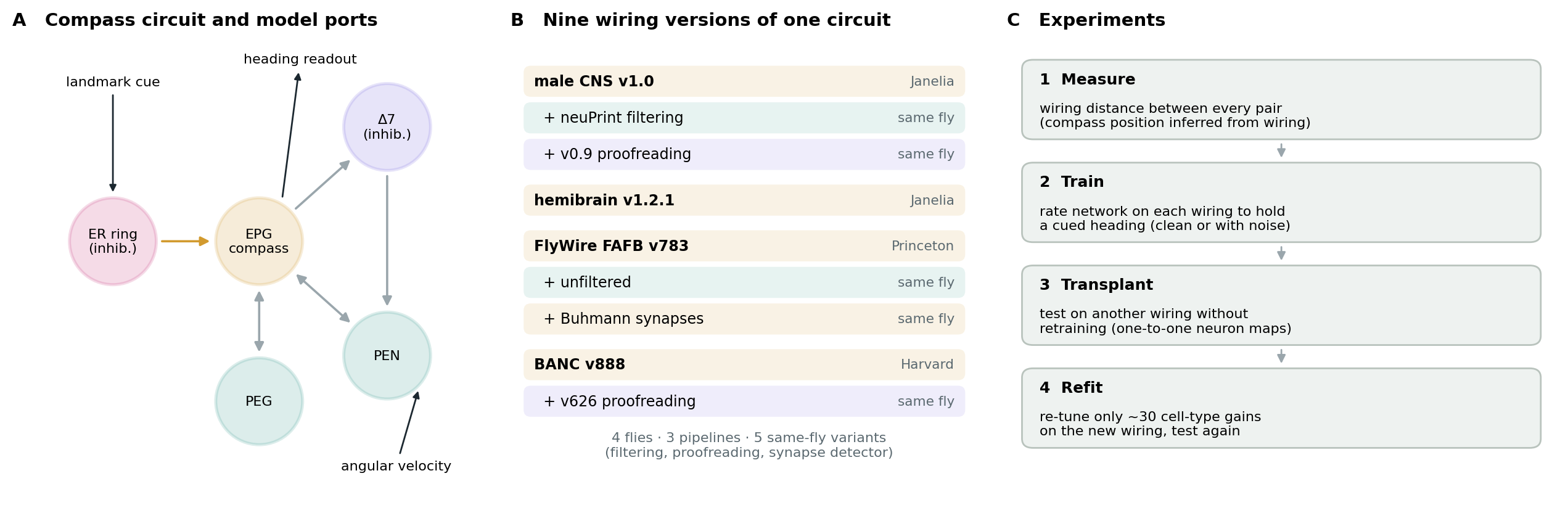}
\caption{\textbf{Study design.} (\textbf{A})~The compass circuit and model ports are: cue to ER, angular velocity to PEN, readout from EPG.  (\textbf{B})~Nine wiring versions: four flies from three groups (bold) and five same-fly variants, in which the variants differ in filtering, proofreading or detection of synapses. (\textbf{C})~Experiments: Distance measurements, Training on each wiring, Transplant through neuron maps, Cell-type gains.}
\label{fig:1}
\end{figure}

To separate reconstruction from biology, we added five variants in which the fly remained unchanged and only a single step of the reconstruction changed (Fig.~\ref{fig:1}B). For the male CNS, a second filtering pipeline (neuPrint instead of Codex) and an earlier proofreading release (v0.9). For FlyWire, the same synapse table without the Codex filter, and an independent synapse detector \citep{buhmann2021} instead of Princeton synapses \citep{yu2025}. For BANC, an early proofreading version (v626). The male CNS network in Codex was identical to the one used in our previous study \citep{chessfly_v1} on all connections shared between them  ($r = 1.000$), and the rebuilt hemibrain network was identical to the previous hemibrain network.

To compare wiring between flies, one must know the location on the compass of each neuron. This position was inferred from wiring alone: EPG neurons were positioned by multidimensional scaling of their connectivity, and all other neurons were positioned to their EPG partners  (Sec.~\ref{sec:m-positions}). In the two flies with independent glomerulus labels, inferred positions agreed with the labels (circular agreement 0.99--1.00; shuffled controls 0.24--0.55) (Table~\ref{tab:s-positions}).

One-to-one neuron maps were constructed between each pair of flies, using these same positions, and were constructed by rank matching after aligning the two compasses by rotation, and, where necessary, mirroring (Sec.~\ref{sec:m-maps}). In the two labelled flies, the maps assigned 100\% of EPG, PEN\_a and PEN\_b neurons and 83\% of PEG neurons to the correct glomerulus. In all 12 ordered fly pairs, 91--99\% of the source neurons were used in the map (Table~\ref{tab:s-maps}).

\subsection{Measurement rivals individuality}
\label{sec:r2}

We compared versions with a wiring-map distance: the connection strength as a function of the angular offset between the two neurons, computed using a total-variation distance weighted by the percentage of synapses of the pathway  (Sec.~\ref{sec:m-distance}). There are two controls that set the floor. When the same male CNS fly was processed by two different filtering pipelines, it resulted in 0.022, while a random 10\% of connections was removed to yield 0.032 (SD 0.003, five repeats).

\begin{figure}[tbp]
\centering
\includegraphics[width=\linewidth]{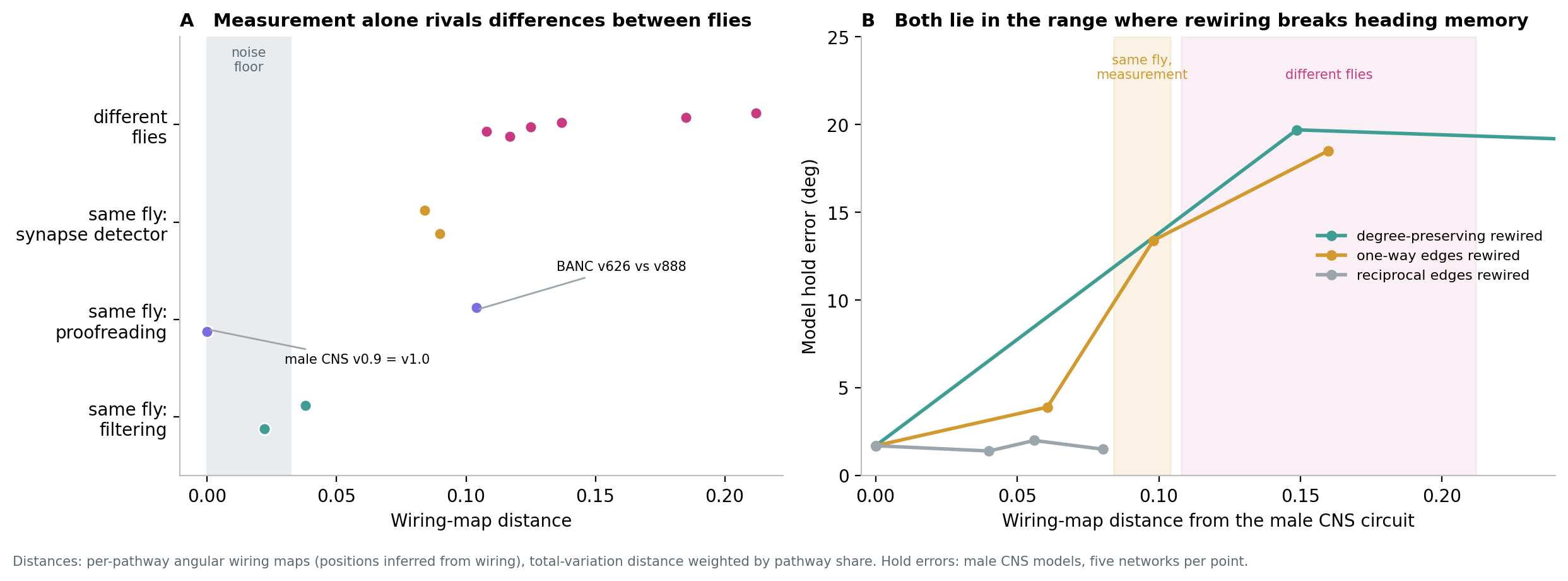}
\caption{\textbf{Connectome differences on one scale.} (\textbf{A})~Wiring-map distance by source of difference; grey band, noise floor (0.022--0.032). (\textbf{B})~Hold error of male CNS models after artificial rewiring \citep{chessfly_v1} against distance from the original (five networks per point); bands mark same-fly measurement and between-fly differences. The degree-preserving curve continues to the 50\% shuffle (0.42, 18.2$^\circ$), off the plotted range.}
\label{fig:2}
\end{figure}

Changing how synapses are detected in a single fly moved the wiring a large part of the way toward a change of fly (Fig.~\ref{fig:2}A). The two comparisons of detector results on FlyWire yielded 0.084 and 0.090. Distances between flies ranged from 0.108 to 0.212 (Table~\ref{tab:distances}). The detector distances were therefore 78--83\% of the smallest between-fly distance, and below the three closest fly pairs (0.108--0.125); the two largest distances (0.185 and 0.212) both involve BANC.

In one fly, proofreading was important and in the other, it was not. For this circuit at distance 0.000, the two male CNS releases (v0.9 and v1.0) did not differ since it was completely proofread before v0.9. The difference between the two BANC releases was 0.104, close to the smallest between-fly distance (0.108). Filtering changes remained around the lower limit of the range (0.022--0.038). 

In the between-fly distances, biology and reconstruction are mixed: only one fly pair (male CNS and hemibrain, 0.117) shares a pipeline family. Therefore, the same-fly variants bounded the measurement component: on this circuit, one change of synapse detector creates differences similar to the closest fly-to-fly comparisons. The detector effect was measured in one fly (FlyWire) and the Buhmann detector is the older of the two.

The ranking was robust across three other distance measures  ($\rho = 0.77$--$0.93$, $p \leq 0.004$; Table~\ref{tab:s-altmetrics}). Differences between detectors were 56--115\% of the nearest between-fly difference depending on the measure and the detector comparison. The BANC proofreading difference ranged from 15\% of the closest between-fly difference for the weights of the neurons to 96\% for the angular map.

Left and right halves of one fly differed no more than random halves of the same neurons: 0.060 versus 0.068 in male CNS and 0.100 versus 0.111 in BANC (five repeats for each side per circuit). At this resolution we detect no bilateral asymmetry.

These distances can be measured on a rewiring-scale of a trained model. Moving 10\% and 20\% of one-way connections raised hold error to 13.4$^\circ$ and 18.5$^\circ$, and a 25\% degree-preserving shuffle to 19.7$^\circ$, while moving 20\% of reciprocal connections did not break holding (1.5$^\circ$; \citealp{chessfly_v1}). On the wiring-map scale these manipulations lie at 0.098, 0.160, 0.149 and 0.080 (Fig.~\ref{fig:2}B). Same-fly measurement differences and between-fly differences both lie within this range, where artificial rewiring breaks holding.

\subsection{Trained models are knife-edge ring attractors}
\label{sec:r3}

For each of the four real wirings we trained a rate network to store a cued heading, either with one gain per neuron or one gain per cell type, and a different seed each time  (Sec.~\ref{sec:m-model}). Of 40 networks, 36 trained; seed 2 failed under both parameterizations in the male CNS and the hemibrain and was excluded by the registered rule (own error $>80^\circ$). The error of the trained networks was 0.3--1.6$^\circ$ after 40 steps.

\begin{figure}[tbp]
\centering
\includegraphics[width=\linewidth]{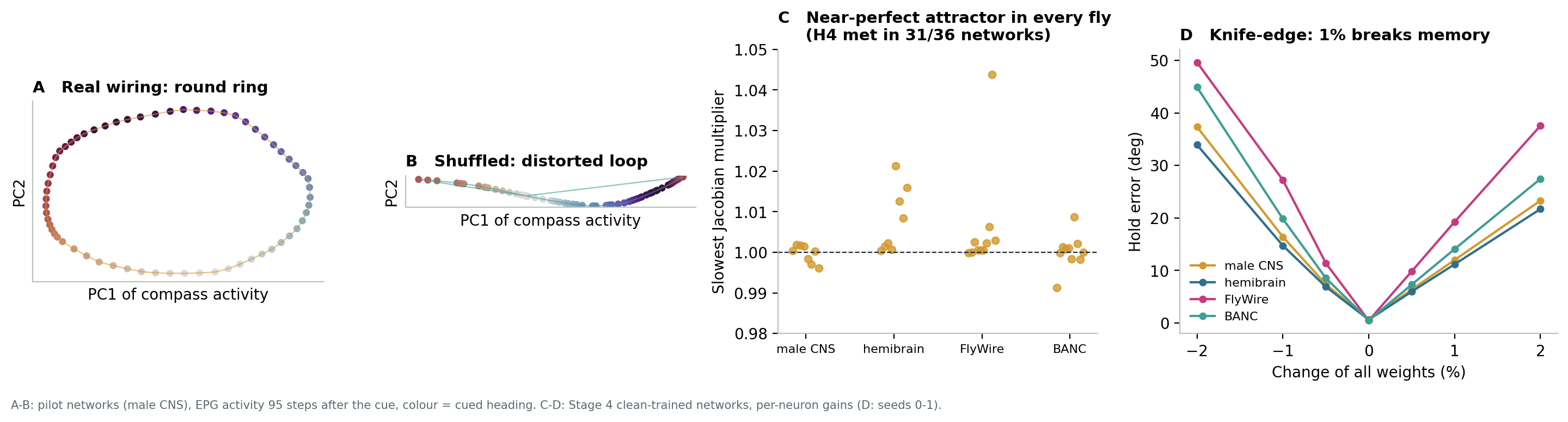}
\caption{\textbf{Near-perfect but knife-edge ring attractors.} (\textbf{A, B})~EPG activity of pilot networks 95 steps after the cue in the top two principal components, coloured by cued heading (same scale): round ring for real wiring, distorted loop for maximally shuffled wiring. (\textbf{C})~Slowest Jacobian multiplier of every trained Stage~4 network; dashed line, 1. (\textbf{D})~Hold error after scaling all weights by $-2$ to $+2\%$ (per-neuron gains, median of seeds 0--1).}
\label{fig:3}
\end{figure}

All real wirings generated a ring attractor with the pre-registered criteria (H4; Fig.~\ref{fig:3}C). Thirty-one of the 36 networks had a ring in the top two principal components of EPG activity and returned to it at a perturbation rate greater than 5 times the rate at which they wandered along it. In all eight fly $\times$ parameterization combinations (3--5 networks each; Table~\ref{tab:s-stats}), the registered rule, at least three of five seeds, was satisfied. Activity had no discernable decay or growth along the ring; the median of the slowest of the Jacobian multipliers of the trained two step map was 0.998–1.014 for each combination, with individual network multipliers between 0.991 and 1.044: a near-continuous attractor.

The criteria detected the positive control. Without retraining, silencing $\Delta$7 neurons resulted in breaking H4 in 26/31 networks and increased hold error in all 31 networks.

Shape separates the real wiring from rewired wiring. In the pilot (male CNS, Supplement S1), real-wiring networks held heading on a round ring (radius variation 0.17; Fig.~\ref{fig:3}A). Networks trained on maximally shuffled wiring also held heading but stored it on a distorted loop (radius variation 0.50--0.54; Fig.~\ref{fig:3}B) that persistent homology did not classify as a ring. In the 12 shuffled pilot networks (25\% and maximal shuffles), the slowest multiplier was 0.900--0.997, with a large gap to the next, indicating an approximate rather than near-continuous attractor.

The nearly perfect attractors were fragile (Fig.~\ref{fig:3}D). All networks tested (two seeds per fly) were extremely sensitive to scaling, with hold error increasing to 10.3--27.9$^\circ$ with a 1\% increase or decrease of all weights, and to 20.5--50.9$^\circ$ with a 2\% change. We take this to mean the fine tuning problem of continuous attractors \citep{seung1996,sagodi2024}: a multiplier close to 1 needs to be tuned, and a uniform change of the weights takes it away from 1.

The same effect was seen when real weak connections were added. Male CNS networks tested on the neuPrint version of the same fly, which keeps weak connections that the Codex filter removes, reached 13--74$^\circ$ (13--38$^\circ$ with cell-type gains, 31--73$^\circ$ with per-neuron gains), although the two versions are separated by only the noise-floor distance of 0.022.

\subsection{Models do not transfer between connectomes}
\label{sec:r4}

\begin{figure}[tbp]
\centering
\includegraphics[width=\linewidth]{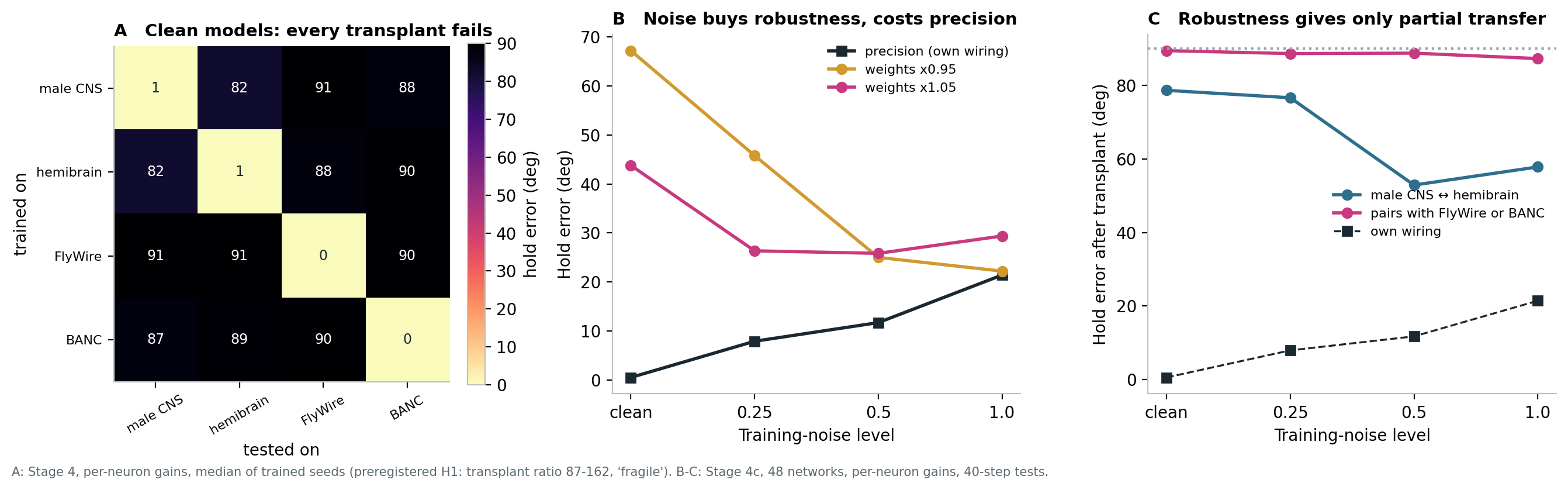}
\caption{\textbf{No transfer; noise trades precision for robustness.} (\textbf{A})~Error of clean per-neuron networks tested on each wiring (rows: trained on; median of trained seeds). (\textbf{B})~Exploratory noise sweep: own-wiring error and error with all weights scaled by 0.95 or 1.05, by training-noise level. Chance, dotted line. (\textbf{C})~Transplant error for male CNS $\leftrightarrow$ hemibrain and for pairs with FlyWire or BANC, with own-wiring error; dotted line, chance.}
\label{fig:4}
\end{figure}

Given this fragility, zero-shot transplants were expected to fail; we tested them as preregistered. To test each trained network, we evaluated it on each other fly's wiring without retraining, carrying its parameters over through the one-to-one neuron maps (Sec.~\ref{sec:m-transplant}). All transplants failed  (Fig.~\ref{fig:4}A). The median error for the ordered fly pairs was 82--92$^\circ$, compared with 0.3--1.6$^\circ$ for the networks' own wiring, and 90$^\circ$ for chance, with 101 of 108 individual transplants being larger than 80$^\circ$, which was the registered failure level.

For the preregistered hypotheses: H1 made two predictions of the transplant ratio $R$, the median error on the target divided by the median error on the source. The fragile prediction (median fly pair with $R > 5$) was supported (median $R = 87$ for cell-type gains, the registered parameterization, and $R = 162$ for per-neuron gains). H2 was supported: neuron-level $R$ was larger than cell-type $R$ in all 12 ordered pairs (one-sided Wilcoxon $p = 2.4\times10^{-4}$, Holm-corrected $7.3\times10^{-4}$). H3, that transplant failure increases with wiring distance, was not supported ($\rho = -0.33$ for cell-type gains, $p = 0.85$; $\rho = +0.25$ for per-neuron gains, $p = 0.21$). Since each transplant was already close to chance, there was no gradient for distance to predict (floor effect).

There are three controls to eliminate procedural explanations. First, mapping each network onto its own wiring via the same rank-matching process reproduced its own-wiring performance, so the mapping does not explain the failures. Second, the maps were accurate for EPG and PEN neurons in the labelled flies (Sec.~\ref{sec:r1}). Thirdly, the failure does not need a different fly: networks failed on the neuPrint version of their own fly (Sec.~\ref{sec:r3}).

One global gain failed to save the transplants. For the seed 0 networks (clean and noise level 0.5; 24 transplants over 12 ordered pairs) we tried a search for the best uniform scaling of target weights between 0.8 and 1.2. The median error was only reduced from 87$^\circ$ to 83$^\circ$ (one-sided Wilcoxon signed-rank test, $p = 6.3\times10^{-5}$, exploratory) and the best scaling was frequently at the end of the range searched for. So transfer failure is not due to overall weight scale.

Training with noise introduced robustness with a tradeoff of precision (Fig.~\ref{fig:4}B). To explore this, we have trained networks on four levels of weight-scale jitter, synaptic noise, dropout noise and activity noise (0, 0.25, 0.5 and 1.0; three seeds per level). Excluding the two networks that did not manage to converge in the absence of any noise, the own-wiring error rose as noise increased from 0.6$^\circ$ to 7.9, 11.7 and 21.4$^\circ$ (medians; $\rho = +0.83$), and the error after scaling all weights by 0.95 decreased from 67$^\circ$ to 46, 25 and 22$^\circ$ ($\rho = -0.86$; both $p < 10^{-11}$).

Robustness gave only partial transfer (Fig.~\ref{fig:4}C). The median error from the transplant was 79$^\circ$ in clean networks, 53$^\circ$ at noise level 0.5 (range 46--81$^\circ$, six networks) and 58$^\circ$ at level 1.0, still far above the 11.7$^\circ$ that these networks achieved when their wiring was not distorted. All pairs with FlyWire or BANC remained close to chance at all noise levels (medians 87-90$^\circ$).

\subsection{Transfer fails with distance, whatever its source}
\label{sec:r5}

To learn which differences a model can absorb, we used seven wiring versions, the four flies and three same-fly variants (male CNS neuPrint, FlyWire unfiltered, FlyWire Buhmann), and trained networks with noise level 0.5 on each of these wiring versions, and transplanted each network to the other six, for a total of 21 networks, 42 ordered pairs, and 126 transplants. For each zero-shot transplant, one gain offset per target cell type was refitted - approximately 30 numbers, while the other parameters were fixed (Sec.~\ref{sec:m-transplant}). These analyses were not pre-registered.

\begin{figure}[tbp]
\centering
\includegraphics[width=\linewidth]{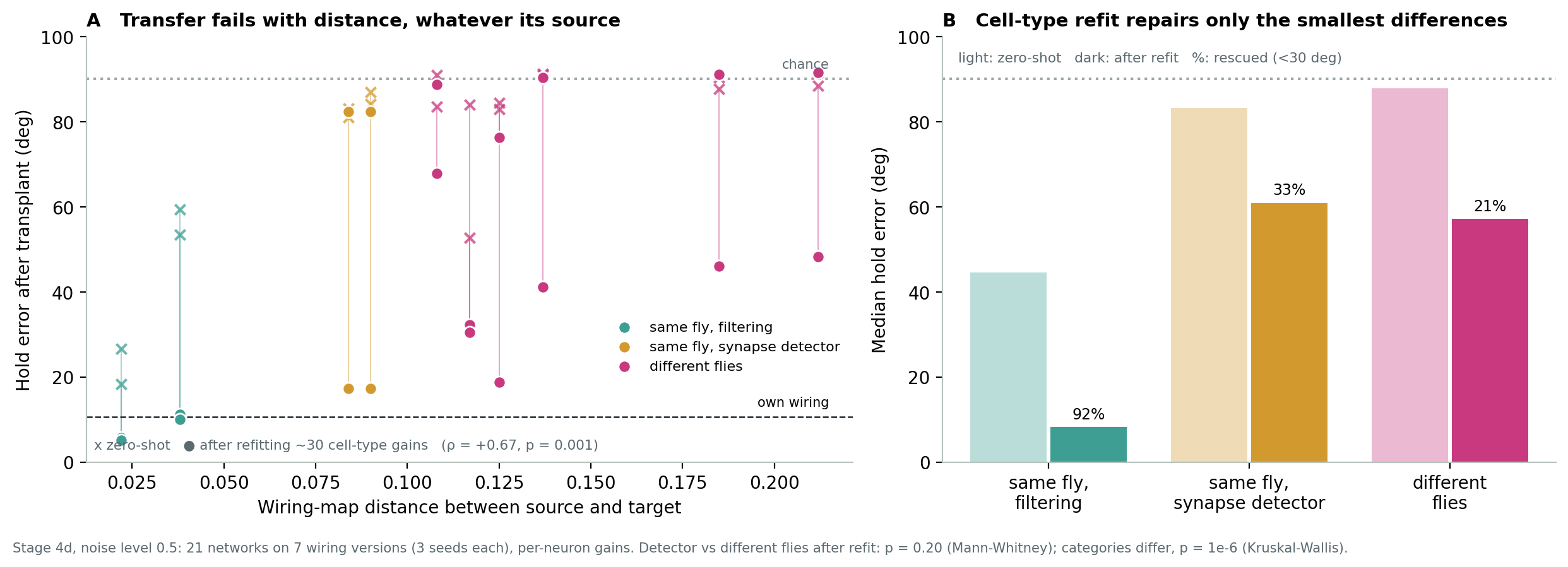}
\caption{\textbf{Transfer depends on distance, not on the source of the difference.} (\textbf{A})~Transplant error against source--target distance, zero-shot (crosses) and after refitting about 30 cell-type gains (dots); dashed line, median own-wiring error over the plotted pairs. (\textbf{B})~Median error by source; percentages, transplants below 30$^\circ$ after refitting. Exploratory: 21 networks on seven versions, noise level 0.5; three refits did not complete.}
\label{fig:5}
\end{figure}

Refitting corrected filtering-level but not detector-level or fly-level differences (Fig.~\ref{fig:5}B). The error of same-fly filtering variants decreased from a median of 45$^\circ$ zero-shot to 8$^\circ$ after refitting, and 92\% of transplants ended below 30$^\circ$. With a different synapse detector, error fell from 83$^\circ$ to 61$^\circ$ (33\% below 30$^\circ$), and with a different fly from 88$^\circ$ to 57$^\circ$ (21\%). The three categories differed (Kruskal--Wallis $p = 10^{-6}$), but after refitting the detector change and the change of fly did not differ (61$^\circ$ against 57$^\circ$, two-sided Mann--Whitney $p = 0.20$).

Refit error increased with the wiring distance between source and target (Fig.~\ref{fig:5}A; Spearman $\rho = +0.67$, $p = 0.001$, 20 ordered pairs with a measured distance). At the same distance, a detector change within one fly was about as damaging as a change of fly. Clean-trained networks showed the same pattern at lower rescue rates: after refitting, 60\% of filtering-level transplants fell below 30$^\circ$, against 17\% of detector-level and 5\% of between-fly transplants.

\subsection{The fly compass degrades gradually}
\label{sec:r6}

\begin{figure}[tbp]
\centering
\includegraphics[width=\linewidth]{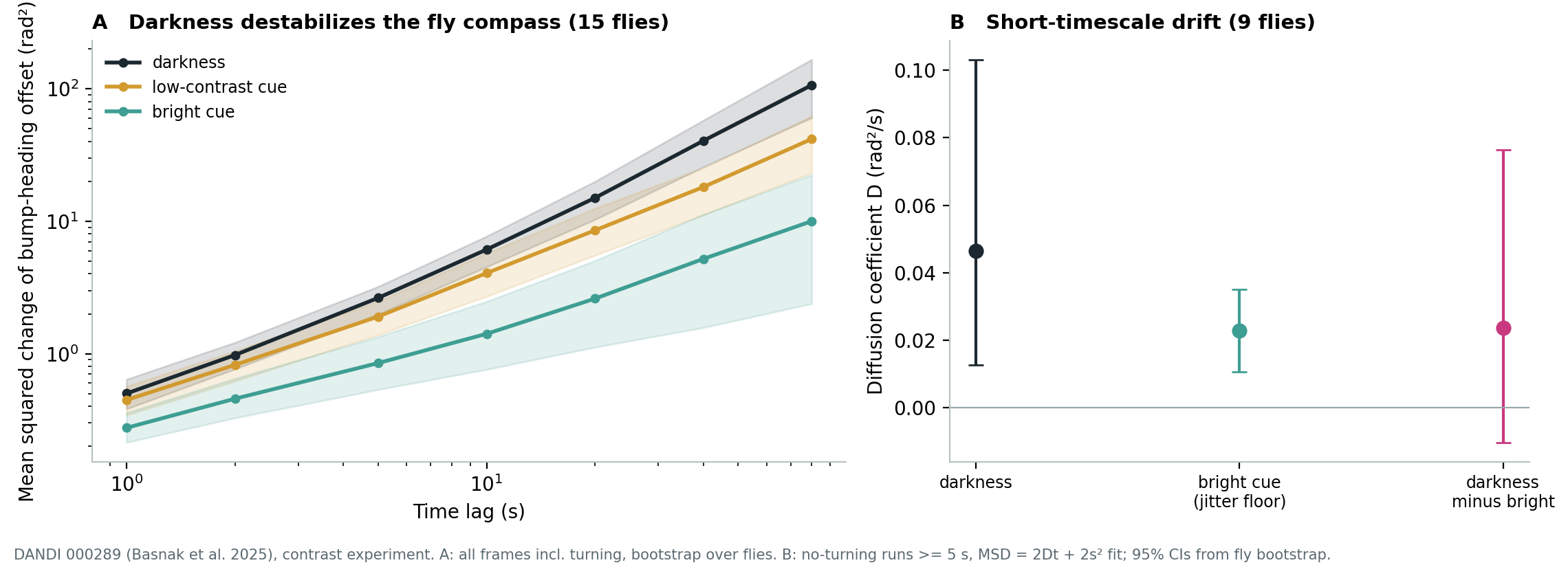}
\caption{\textbf{The fly compass degrades gradually.} (\textbf{A}) Mean squared change of bump-heading offset against lag (15 flies, all frames; shading, 95\% CI over flies). (\textbf{B})~Diffusion coefficient from $\mathrm{MSD} = 2Dt + 2\sigma^2$ fits to no-turning runs of at least 5~s (9 flies); bars, 95\% CI. Data: DANDI 000289 \citep{basnak2025}.}
\label{fig:6}
\end{figure}

We examined E-PG calcium imaging from 15 flies in the presence of a bright visual cue, a low-contrast cue, and darkness (DANDI 000289; \citealp{basnak2025}). In all three conditions the bump--heading offset spread with time lag, fastest in darkness (Fig.~\ref{fig:6}A). At a lag of 40~s, the mean squared difference was 40.4~rad$^2$ in darkness (95\% CI 25.6--57.7, bootstrap over flies), 18.1~rad$^2$ with a low-contrast cue and 5.2~rad$^2$ with a bright cue (1.6--11.2), about eight times smaller than in darkness, with non-overlapping intervals.

On short timescales we did not establish the excess drift that darkness would contribute. A fit to the mean squared change on no-turning runs over a minimum 5~s for the 9 flies with data in both conditions (Sec.~\ref{sec:m-drift}) gives $D = 0.047$~rad$^2$/s in darkness (0.013--0.103) and 0.023~rad$^2$/s with a bright cue (0.011--0.035), the measurement jitter floor. The difference, 0.024~rad$^2$/s ($-0.010$ to 0.076), includes 0. We did not compare these with drift in the models (Sec.~\ref{sec:m-prereg}).

% ---- Tables referenced in Results (main text) ----
\begin{table}[tbp]
\centering
\small
\caption{\textbf{The nine wiring versions.} Head-direction circuit (EPG, PEG, PEN, $\Delta$7, ER1--6), at least three synapses per neuron pair.}
\label{tab:versions}
\begin{tabular}{@{}llrr@{}}
\toprule
Version & Source & Neurons & Connections \\
\midrule
male CNS v1.0 & FlyWire Codex (filtered) & 434 & 27,246 \\
\quad male CNS, neuPrint & neuPrint & 434 & 33,783 \\
\quad male CNS v0.9 & Codex & 434 & 27,246 \\
hemibrain v1.2.1 & neuPrint & 409 & 28,942 \\
FlyWire FAFB v783 & Codex, Princeton synapses, filtered & 433 & 16,741 \\
\quad FAFB unfiltered & Codex unfiltered file, $\geq$3 synapses & 433 & 23,340 \\
\quad FAFB Buhmann & Codex, Buhmann synapses & 433 & 18,452 \\
BANC v888 & Codex & 419 & 15,440 \\
\quad BANC v626 & Codex & 419 & 17,343 \\
\bottomrule
\end{tabular}
\end{table}

% ===================== DISCUSSION =====================
% Citation keys are provisional; every reference is opened and verified in step 8.
\section{Discussion}
\label{sec:discussion}

\subsection{Measurement is a large part of connectome variability}

On the fly compass, reconstruction contributed substantially to connectome differences. Replacing only the synapse detector within one fly produced more than half of the smallest between-fly difference on all four measures. Proofreading differences were smaller and measure-dependent. Because only one pair of flies is processed by the same pipeline (the other fly pairs differ in their reconstruction pipeline), the between-fly distances are not entirely reconstruction. We can only say that the measurement component is large enough that a connectome difference cannot be attributed to individuality without a same-fly comparison. 

This accords with prior evidence that synapse-detection performance varies between data sets \citep{scheffer2025}, that many FlyEM--FlyWire differences are technical \citep{okuno2025}, and that setting dataset-specific thresholds helps secure reliable connections \citep{berg2026}. Here we contribute a controlled comparison: same-fly reconstructions that differ by one step at the same scale as fly-to-fly differences, combined with a functional assessment of how those differences affect a trained model.

\subsection{Why trained models fail: the fine-tuning problem}

The fragile state of the trained networks is a consequence of the solution that training leads to. Holding a heading without input requires a continuum of stable states, which the rate network obtains by setting its slowest mode to a multiplier of 1 (close to what our networks achieved, median 0.998-1.014 per fly and parameterization). This is a structurally unstable solution: any change to the weights causes the multiplier to deviate from 1, and the bump to drift or vanish \citep{seung1996}.

Thus, transplant failure is not a sign of incompatibility among individual fly circuits, but rather a consequence of their inherent fragility. Any network that cannot adapt to its weights changing by 1\% will fail for any connectome that is different by more than that, and all the ones we tested did, including other reconstructions of the same fly. That's also why we failed to support H3: at chance levels for each transplant, there was nothing left for distance to predict, whereas after noise training and refitting, error increased with distance.

That same reasoning points to a solution. Perturbed continuous attractors typically depart the slow manifold that held the variable for some time \citep{sagodi2024}, and hidden symmetries in the heterogeneous compass connectivity can support ring-attractor dynamics \citep{hulse2026}. But approximate attractor shape alone was not sufficient in our hands: shuffled-wiring pilot networks stored heading on a deformed loop (multipliers 0.900-0.997), and were as sensitive to 1\% weights change as real-wiring networks (Table~\ref{tab:s-pilotscale}). Noise training conferred robustness at the price of accuracy. A real circuit model may need to be trained to become robust, rather than left at the knife-edge produced by clean training.

In flies, the bump--heading offset spread about eight times faster in darkness than with a bright cue at a 40~s lag, consistent with a slowly drifting approximate attractor that sensory input corrects. Which mechanisms would make trained models robust, such as developmental noise, plasticity or feedback from other circuits, is a question for future work.

\subsection{Relation to static fits of the four connectomes}

\citet{biswas2024} demonstrated that all four connectomes could be mapped into the ring attractor after a rescaling of a handful of cell types. Our result is not incompatible with that: these two results are answering different questions. Static fit simply means independently fit each connectome to an ideal model. Transplant tests whether a parameter set trained on one connectome generalizes to another, without refit from scratch. If we allow refitting at only roughly 30 cell-type gains, then we can at least correct for filtering-level, but not detector-level or fly-to-fly differences. That there are only a handful of cell-type parameters that make each connectome an attractor does not mean that those same parameters, taken from one connectome's solution, can build a trained model for another. This fits the degeneracy of connectome-constrained models \citep{beiran2025}: many parameter configurations will elicit the target dynamics when limited to a single wiring. The parameter configuration a model that encounters one wiring ultimately converges to will not necessarily be close to any good set of parameters when limited to the other wiring.

\subsection{Alternative explanations}

The transfer did not fail because of neuron mismatch or overall weight scale. The maps were accurate for the EPG and PEN neurons on the labelled flies, self-mapping replicated own success, networks failed on a different reconstruction of own fly where no cross-fly alignment is employed, and the best global gain improved transplants only from  87$^\circ$ to 83$^\circ$. The discrepancy is in the pattern of connections, not their magnitude.

The difference between a round ring and a distorted loop is not a matter of visual judgement: the ring criteria were preregistered, and Jacobian multipliers, winding number and radius variation are all provided for every network (Tables~\ref{tab:s-pilotmech} and~\ref{tab:s-h4}; the last two are exploratory).

The precision-robustness trade-off applies to one particular class of rescaled noise models; other types of noise and the deterministic sensitivity penalty may outperform.

\subsection{What modelers should do}

First, include the original reconstruction: dataset, version, synapse table, connection threshold and proofreading release. For this circuit, two synapse tables of one fly differed by more than half as much as the two most similar flies, and two proofreading releases by up to 96\% on one measure.

Second, a model trained on one connectome should be a separate entity until it is reproduced on the other, and parameters should be tuned for each set separately. Cell-type parameters moved between filtering variants of a fly, but not between synapse detectors or flies.

Third, always test the trained model for robustness. Before you can interpret what a trained model says about a given circuit, two trivial checks are whether the error increases after scaling all weights by 1\% and whether the error increases on a second reconstructed version of the same circuit (preferably of the same animal).

Fourth, consider same-fly variants as the reference before ascribing a difference between connectomes to individuality. Compass circuit: the measurement floor set by filtering is small (0.022-0.038 in our scale), while the measurement floor set by synapse detection and proofreading is not.

\subsection{Limitations}

Our data includes four flies reconstructed by three groups. Biology and pipeline are integrated with this design and the same-fly variants limit the measurement component but do not eliminate it from the between-fly distances.

Between-fly distances at the level of individual neurons are upper bounds: within a fly, neurons are matched by identity, but between flies they are matched by inferred maps, which contribute residual error to the measured difference.

The measurement effects were estimated in only two flies: the synapse-detector effect in FlyWire and the proofreading effect in BANC. For the male CNS, proofreading of this circuit had been finished in the earlier release.

Our models are rate networks that learned with backpropagation on a single task, with a cued heading, in the absence of angular-velocity input. We did not test angular integration or lateralized input, although both were preregistered (Sec.~\ref{sec:m-prereg}). Different classes of model (e.g., spiking, fits to activity data) might be more or less robust.

It was not possible to establish a short-time-scale excess drift of the fly compass in darkness for nine flies, nor was it possible to determine if there was a distinction between fly drift and model drift because the networks we trained are deterministic and the available model reference value has an unverified variance convention. The noise sweep, cell-type refit and same-fly transfer analyses were not preregistered; they are exploratory, and the reported statistics are uncorrected.

\subsection{Future work}

More same-fly reconstructions, with multiple synapse detectors or proofreading releases of a single animal, would turn the measurement bound into an estimate, and more flies from a single pipeline would do the same for individuality. Robustness training, a registered integration task with lateralized input as done by  \citet{duan2025}, would test whether robust models generalize where fragile ones do not. Connectomes combined with activity recordings from the same animal would enable comparison of trained models with the drift of their own circuit.

\subsection{Conclusion}

Reconstructions of the compass circuit of one fly can vary in a significant fraction of the variation between flies, and networks trained on them are close to perfect knife-edge ring attractors that fail under either type of difference. Connectome-constrained models should report the reconstruction they were built on and be tested for robustness across reconstructions.

% ===================== METHODS =====================
\section{Methods}
\label{sec:methods}

\subsection{Connectome extraction}
\label{sec:m-extraction}

Male CNS v1.0, FlyWire FAFB v783 and BANC v888 were downloaded from FlyWire Codex on 14 October 2026 using the filtered connection files. Hemibrain v1.2.1 was queried from neuPrint  \citep{plaza2022}. Cell types were selected based on matching names with the regular expression \texttt{\^{}(EPG|PEG|PEN|Delta7|ER[1-6])}, and their all-to-all connectivity matrices summed over neuropils, connecting pairs with fewer than 3 synapses removed, and self-connections eliminated. $\Delta$7 and ER class neurons were set to inhibitory (negative) weights, other neurons to excitatory (positive) weights. Cell types were collapsed to families for mapping (EPG, EPGt, PEN\_a, PEN\_b, PEG, $\Delta$7 and the ER subtypes ER1, ER2, ER3a, ER3d, ER3m, ER3p, ER3w, ER4d, ER4m, ER5, ER6). To test robustness to proofreading version and synapse detection, we built alternative networks from the same animals with the same rules: for the male CNS, the neuPrint network from our previous paper and the Codex v0.9 release; for FAFB, the Codex unfiltered connection file (Princeton synapses) and the connection file based on synapses predicted with the detector of \citet{buhmann2021}, both constrained to the same 433 neurons selected from v783; and for BANC, the v626 release. The Codex male CNS network had identical synapse numbers as our previous paper's neuPrint network for every connection present in both  ($r = 1.000$), although Codex missed approximately 19\% of neuPrint connections. The rebuilt hemibrain network was identical to our previous paper's.

\subsection{Compass positions and neuron maps}
\label{sec:m-positions}
\label{sec:m-maps}

Each EPG's concatenated input and output vector, normalized to unit length, was fed into classical multidimensional scaling of the cosine-similarity matrix; its angle was given by the first two axes. All other neurons were assigned at the synapse-weighted circular mean of its EPG partners' angle. For the two flies with glomerulus labels (male CNS community labels, hemibrain instances), the per-side circular agreement $|\langle e^{i(\phi_j - k\lambda_j)}\rangle|$ between inferred angles $\phi_j$ and label angles $\lambda_j = 2\pi(g_j-1)/8$,was averaged over labelled neurons and maximized over  $k = \pm1$, with label shuffles as controls.

To align each ordered fly pair, neurons were matched within each family by rank of angle around the ring (stretched for count differences, with the circular offset chosen to minimize angular differences). The target frame was chosen among 72 rotations, with and without mirroring, maximizing the $\log(1+\text{synapse count})$ correlation in direction-tuned cells (all but ER); task performance was not used for aligning. A rank match was used instead of a nearest-angle rule after testing (deviation 5).

\subsection{Wiring-map distance}
\label{sec:m-distance}

For each pathway $p$ of the direction-tuned families, (EPG, PEN a, PEN b, PEG, $\Delta$7), the angular profile $m_p$ is the distribution of edge weight over the angular offset from presynaptic to postsynaptic neuron, in 8 bins of 45$^\circ$, normalized to sum to 1. The distance between version  $a$ and $b$ is
\begin{equation}
d(a,b) = \sum_{p} w_p \,\tfrac{1}{2}\| m^a_p - m^b_p \|_1 \Big/ \sum_p w_p ,
\end{equation}
where $w_p$ is the average over both versions of the fraction of the postsynaptic family's input (over all families, including ER) that comes from the presynaptic family. Normalization controls for differences in total number of synapses (142,000-538,000 in this circuit). ER neurons are not direction-tuned, so ER pathways do not contribute any profile and were excluded. Angles are only defined up to reflection, so each version was oriented once to best match the male CNS v1.0 maps, then all distances were measured between these oriented maps. Controls: male CNS from two pipelines (Codex, neuPrint), male CNS with 10\% of connections dropped at random, and left versus right halves versus random halves (5 repeats each). Rewired circuits from our previous paper \citep{chessfly_v1} were measured the same way.

To evaluate robustness to measure choice, each contrast was also scored with three generic measures: 1 minus the Spearman correlation of neuron-level input fractions over the union of edges, the Jaccard distance between edge sets, and the cell-type pathway-share distance. For the first two we used only direction-tuned neurons (EPG, PEN, PEG, $\Delta$7), with data matched by neuron identity within a fly and by the rank map across flies. ER neurons are not used in these comparisons because their weak tuning prevents consistent matching across flies. 

\subsection{Network model, task and training}
\label{sec:m-model}
\label{sec:m-training}

The model was adapted from our previous study \citep{chessfly_v1}: the state $h$ is being updated twice per one time step
\begin{equation}
h \leftarrow \tfrac{1}{2}h + \tfrac{1}{2}\left[\mathrm{ReLU}(h)\,W + u_t\right], \qquad W_{ij} = s_i\, e^{\gamma_i}\,\kappa\, C_{ij},
\end{equation}
with synapse counts $C$ (rows presynaptic), signs $s_i$, $\kappa$ scaling $\mathrm{diag}(s)C$ to spectral radius 0.9, and trainable log-gains  $\gamma_i$, one per neuron or one per cell type. For the first 5 steps the cue  $(\cos\theta, \sin\theta, 1)$ feeds the ER neurons via learned weights and biases; afterwards ER neurons receive only their biases. PEN neurons each receive a learned bias; they would receive an angular-velocity input, but in the hold task it is zero. Heading is read out as $\hat\theta = \mathrm{atan2}(\hat y_2,\hat y_1)$ with $\hat y = \mathrm{ReLU}(h_{\mathrm{EPG}})R + r$.  Error is average magnitude of wrapped difference between  $\hat\theta$ and $\theta$ after the cue; chance is 90$^\circ$, and networks with own-wiring error above 80$^\circ$ were counted as failed.

The networks were trained in PyTorch  \citep{paszke2019} using Adam  \citep{kingma2015} (learning rate $10^{-3}$, batch 128, 8,000 updates) on the MSE between  $\hat y$ and $(\cos\theta,\sin\theta)$,with per-network gradient clipping at norm 1.0 and the best-validation parameters (checked every 250 updates) kept. A total of 1,000, 500 and 2,000 runs were used for training, validation and testing respectively (fixed seeds per network, log-gains init by zero). Networks were padded to 434 neurons and trained in parallel on a single NVIDIA T4 GPU. Noise training was applied at each update with per-network weight scale $\mathcal{N}(1, 0.03^2)$, log-normal SD 0.10 synaptic noise, 5\% synapse dropout, and SD 0.02 activity noise, multiplied by a noise level (0, 0.25, 0.5, 1.0); evaluation was noise-free, and networks with the same seed shared initializations and minibatches across noise levels.

\subsection{Transplants and refitting}
\label{sec:m-transplant}

A transplanted network assigns each target neuron the log-gain, cue weights, biases, readout weight of its mapped source neuron, with target's own counts, signs, $\kappa$, and is tested without further training. $R$, the transplant ratio, is the median target error divided by the median own-wiring error per ordered fly pair, for all networks except those that failed. In global-gain testing, all target weights of the seed-0 networks (clean, 0.5 noise level; 24 transplants) were scaled by the factor in [0.8, 1.2] that minimized error on validation headings, then tested on test headings. In cell-type refitting, one log-gain offset (about 30 numbers) per target cell type was trained for 600 updates (Adam, learning rate 0.02, batch size 128) with all else frozen; a transplant was counted as rescued if below 30$^\circ$.

\subsection{Mechanism measures}
\label{sec:m-mechanism}

Networks were re-simulated on the CPU, which replicated GPU errors exactly (for instance, 4.396$^\circ$ on both for the pilot network). Each was cued with 64 headings, ran for 200 steps. The preregistered ring criterion: the top two principal components of EPG rates 95 steps after the cue together explain at least 70\% of variance and persistent homology  \citep{tralie2018} show 1 dominant feature (top persistence >0.3, at least 3$\times$ the second). The return criterion: $\log\lambda_2/\log\lambda_1 \geq 5$ (from the 2 largest eigenvalue moduli of the Jacobian of the 2-update map at the settled states of 8 headings) so activity returns to the ring 5× faster than drifts along it; H4 supported if, for a fly and parameterization, at least 3/5 trained networks (or trained network) met both criteria. The positive control silenced $\Delta$7 neurons by setting their output weights to zero. Winding number and radius variation in top 2 components are exploratory (deviation 3). The weight-scale test scaled weights 0.98-1.02 without retraining.

\subsection{Fly compass drift}
\label{sec:m-drift}

We analysed DANDI 000289's contrast experiment \citep{basnak2025}: E-PG imaging in the protocerebral bridge of 15 flies, bright cue, low-contrast cue, darkness. Bump phase was the population-vector angle in each bridge region of interest relative to the angle map from the original study; bump-heading offset was the unwrapped difference from heading. Long-timescale drift was the mean squared change of the offset at lags of 1--80 s over all frames; short-timescale drift was fitted as $\mathrm{MSD}(t) = 2Dt + 2\sigma^2$ (lags up to 3 s) to the pooled bump phase in runs of at least 5~s with heading change < 10$^\circ$/s; nine flies had such runs in darkness and with the bright cue, whose $D$ is the jitter floor. Confidence intervals are based on 2,000 bootstrap resamples of flies. When applied to simulated data with known $D$, this process estimates $D$ within $-5.2\%$ and $+2.2\%$ (Supplement S7).

\subsection{Statistics}
\label{sec:m-stats}

The analysis unit is the trained network, with seeds nested in wiring versions. Prespecified directional tests are one-sided: Wilcoxon signed-rank over 12 ordered fly pairs (H2), Spearman correlations (H3), Holm-corrected \citep{holm1979} within the family; H1 and H4 are decision rules on medians and counts. Exploratory analyses are Spearman correlations, Kruskal-Wallis, and two-sided Mann-Whitney tests, and a one-sided Wilcoxon signed-rank test for the global gain, without correction. All tests are listed in Table~\ref{tab:s-stats}.

\subsection{Preregistration and deviations}
\label{sec:m-prereg}

The hypotheses H1-H5, the mechanism criteria, the exclusion rule and the analysis plan were made publicly available on Zenodo on 15 October 2026 (doi:10.5281/zenodo.23382486; \citealp{prereg2026}). At registration, the extraction of the 4 fly circuits, position inference and validation, all between-fly wiring-map distances with the noise floor, as well as the exploratory drift estimate from DANDI 000289 had been completed; none of the networks had been trained on the hemibrain, FlyWire or BANC wiring, none of the transplant or mechanism analysis had been run. The deviations are:

\begin{enumerate}[leftmargin=*]
\item Training for all networks, including stage 4, was increased from 2k to 8k updates because the pilot networks trained at 2k were undertrained.
\item Seed 2 failed to train under both parameterizations in the male CNS and the hemibrain, in the pilot and in Stage 4 and was thus pruned by registered rule.
\item Winding number and radius variation have been incorporated to the H4 measures, due to the fact that persistent homology does not detect distorted loops; these are classified as exploratory.
\item The noise training, noise sweep, same-fly variants, cell-type refit and global-gain test are not registered and are exploratory.
\item The neuron maps were altered from nearest-angle matching to one-to-one rank matching when they were validated against the glomerulus labels, before we ran any of the transplants.
\item The angular integration task, as well as the lateralized angular-velocity input task, were registered but not run; only the hold task with tied input was run.
\item Nulls for the male CNS registered for the new design (degree-preserving shuffles of 25\%, 60\% and maximum, null for cell-types, null for 10\% and 20\% one-way and 20\% reciprocal rewiring) were not run beyond the pilot. The rewiring for our previous study \citep{chessfly_v1} was used for Fig.~\ref{fig:2} .
\item We did not make the H5 comparison of model drift with fly drift; the trained networks are deterministic, so they possess no inherent drift for comparison, and the model reference value available has an unverified variance convention. We present fly drift as a descriptive variable and defer the comparison to future work.
\item Holm correction was registered for H2-H4, but used on H2-H3 only because H4 is the decision rule on the counts (without a $p$ value).
\item H3 was registered as an association of wiring distance with transplant error but was tested on the transplant ratio $R$.
\end{enumerate}

\subsection{Data and code availability}
\label{sec:m-data}

Connectomes are publicly available via FlyWire Codex (male CNS v1.0 and v0.9, FAFB v783, BANC v888 and v626, downloaded 14 October 2026) and neuPrint (hemibrain v1.2.1, male CNS). Fly imaging is available via DANDI (dandiset 000289). Code, extracted connectomes, neuron maps and all result tables are available at  \url{https://github.com/donti507/chessfly} (folder \texttt{new\_study/}). The preregistration is publicly available at doi:10.5281/zenodo.23382486. The authors used AI assistants (Anthropic's Claude) for code development, data-analysis support, literature search and drafting. All analyses, results and text were checked and approved by the authors, who take full responsibility for the content. M.A.A.Y. designed the study, built the network model together with N.T.N., ran the experiments and wrote the manuscript; N.T.N. performed the statistical analyses and built the network model together with M.A.A.Y.; both authors reviewed and approved the manuscript.

\bibliographystyle{plainnat}
\bibliography{refs}
\clearpage
\appendix
\renewcommand{\thetable}{S\arabic{table}}\setcounter{table}{0}
\renewcommand{\thefigure}{S\arabic{figure}}\setcounter{figure}{0}
\section*{Supplementary material}
\addcontentsline{toc}{section}{Supplementary material}

\subsection*{S1.\enspace Pilot}
\label{sec:s-pilot}
The pilot used the male CNS Codex network and trained 18 networks in parallel: real wiring, a 25\% degree-preserving shuffle and a maximal shuffle (70\% of the connections moved) with per-neuron and cell-type gains and three seeds for 8,000 updates. One seed (2) of real wiring failed for both parameterizations (88$^\circ$) and was dropped by the registered rule. Hold errors are shown in Table~\ref{tab:s-pilot}. Maximally shuffled wiring held headings as well as real wiring over 40 steps and better than 25\%-shuffled wiring, replicating the non-monotonic shuffle effect in our previous study with new code and network extraction. 

The four trained real networks satisfied both of H4's criteria  (ring and fast return; Table~\ref{tab:s-pilotmech}). In 12 shuffled control networks, there was no ring by the registered persistent-homology test, despite activity remaining in two dimensions (top-two-component variance 0.962-1.000), and 10 of 12 satisfied the fast-return criterion. Silencing $\Delta$7 neurons increased 200-step hold error in all four real networks (from 4.4--17.9$^\circ$ to 33.3--82.5$^\circ$) and disrupted H4 in three of four. Errors increased over all 16 trained pilot networks from median 3.0$^\circ$ to 21.1$^\circ$ in the worst direction when all weights changed by 1\% (Table~\ref{tab:s-pilotscale}); shuffled controls responded as strongly as real networks.

\begin{table}[H]
\centering
\small
\caption{\textbf{Pilot hold errors} (degrees; test sets of 40, 80 and 200 steps).}
\label{tab:s-pilot}
\begin{tabular}{@{}lllrrr@{}}
\toprule
Wiring & Gains & Seed & 40 steps & 80 steps & 200 steps \\
\midrule
real & per neuron & 0 & 0.68 & 1.26 & 4.40 \\
real & per neuron & 1 & 0.57 & 1.39 & 5.31 \\
real & per neuron & 2 & 88.11 & 88.60 & 89.99 (failed) \\
real & cell type & 0 & 1.23 & 3.97 & 17.51 \\
real & cell type & 1 & 1.11 & 3.69 & 17.96 \\
real & cell type & 2 & 88.11 & 88.60 & 89.99 (failed) \\
25\% shuffle & per neuron & 0 & 4.45 & 7.43 & 16.84 \\
25\% shuffle & per neuron & 1 & 4.01 & 7.28 & 17.15 \\
25\% shuffle & per neuron & 2 & 3.60 & 5.93 & 13.54 \\
25\% shuffle & cell type & 0 & 7.44 & 13.12 & 31.02 \\
25\% shuffle & cell type & 1 & 7.50 & 11.44 & 25.30 \\
25\% shuffle & cell type & 2 & 8.90 & 14.72 & 32.88 \\
maximal shuffle & per neuron & 0 & 1.13 & 1.35 & 3.56 \\
maximal shuffle & per neuron & 1 & 1.89 & 2.65 & 6.82 \\
maximal shuffle & per neuron & 2 & 1.79 & 2.53 & 6.17 \\
maximal shuffle & cell type & 0 & 3.71 & 4.50 & 8.96 \\
maximal shuffle & cell type & 1 & 4.85 & 5.60 & 9.33 \\
maximal shuffle & cell type & 2 & 0.97 & 1.59 & 4.23 \\
\bottomrule
\end{tabular}
\end{table}

\begin{table}[H]
\centering
\footnotesize
\caption{\textbf{Pilot mechanism measures.} Error at 200 steps (degrees), variance in the top two principal components of EPG activity, top persistence of the one-dimensional homology feature, the two largest Jacobian multipliers, and the two H4 criteria. Failed networks (seed~2 of the real wiring) are omitted.}
\label{tab:s-pilotmech}
\begin{tabular}{@{}lllrrrrrcc@{}}
\toprule
Wiring & Gains & Seed & Err. & PC2 & H1 & $\lambda_1$ & $\lambda_2$ & Ring & Return \\
\midrule
real & neuron & 0 & 4.4 & 1.000 & 1.34 & 1.000 & 0.986 & yes & yes \\
real, $\Delta$7 silenced & neuron & 0 & 82.5 & 1.000 & 0.51 & 0.975 & 0.949 & yes & no \\
real & neuron & 1 & 5.3 & 1.000 & 1.18 & 1.000 & 0.997 & yes & yes \\
real, $\Delta$7 silenced & neuron & 1 & 79.8 & 1.000 & 0.22 & 0.995 & 0.978 & no & no \\
real & cell type & 0 & 17.3 & 0.999 & 0.97 & 0.998 & 0.981 & yes & yes \\
real, $\Delta$7 silenced & cell type & 0 & 46.3 & 0.998 & 0.72 & 1.028 & 0.982 & yes & yes \\
real & cell type & 1 & 17.9 & 1.000 & 1.58 & 0.999 & 0.976 & yes & yes \\
real, $\Delta$7 silenced & cell type & 1 & 33.3 & 0.999 & 1.42 & 0.995 & 0.989 & yes & no \\
25\% shuffle & neuron & 0 & 16.9 & 0.995 & 0.00 & 0.961 & 0.824 & no & no \\
25\% shuffle & neuron & 1 & 16.6 & 0.992 & 0.00 & 0.967 & 0.763 & no & yes \\
25\% shuffle & neuron & 2 & 15.7 & 0.989 & 0.00 & 0.986 & 0.780 & no & yes \\
25\% shuffle & cell type & 0 & 31.9 & 0.999 & 0.00 & 0.946 & 0.697 & no & yes \\
25\% shuffle & cell type & 1 & 25.3 & 1.000 & 0.00 & 0.969 & 0.665 & no & yes \\
25\% shuffle & cell type & 2 & 32.8 & 0.999 & 0.00 & 0.900 & 0.695 & no & no \\
maximal shuffle & neuron & 0 & 4.6 & 0.996 & 0.00 & 0.993 & 0.689 & no & yes \\
maximal shuffle & neuron & 1 & 6.8 & 0.997 & 0.00 & 0.997 & 0.715 & no & yes \\
maximal shuffle & neuron & 2 & 7.1 & 0.987 & 0.00 & 0.967 & 0.764 & no & yes \\
maximal shuffle & cell type & 0 & 9.3 & 0.996 & 0.00 & 0.948 & 0.652 & no & yes \\
maximal shuffle & cell type & 1 & 10.0 & 0.995 & 0.00 & 0.965 & 0.714 & no & yes \\
maximal shuffle & cell type & 2 & 4.2 & 0.962 & 0.00 & 0.992 & 0.777 & no & yes \\
\bottomrule
\end{tabular}
\end{table}

\begin{table}[H]
\centering
\small
\caption{\textbf{Pilot weight-scale test.} Hold error (degrees, 40 steps) after multiplying all weights by the given factor, without retraining.}
\label{tab:s-pilotscale}
\begin{tabular}{@{}lllrrrrr@{}}
\toprule
Wiring & Gains & Seed & $\times$1 & $\times$0.98 & $\times$0.99 & $\times$1.01 & $\times$1.02 \\
\midrule
real & neuron & 0 & 0.69 & 33.6 & 14.3 & 11.3 & 21.8 \\
real & neuron & 1 & 0.56 & 48.9 & 26.0 & 18.4 & 36.0 \\
real & cell type & 0 & 1.22 & 39.3 & 18.4 & 12.7 & 27.3 \\
real & cell type & 1 & 1.11 & 30.9 & 11.8 & 9.1 & 16.0 \\
25\% shuffle & neuron & 0 & 4.39 & 53.9 & 29.3 & 26.0 & 44.4 \\
25\% shuffle & neuron & 1 & 3.69 & 66.0 & 39.2 & 30.7 & 53.3 \\
25\% shuffle & neuron & 2 & 3.98 & 60.3 & 32.7 & 33.1 & 59.2 \\
25\% shuffle & cell type & 0 & 7.49 & 53.5 & 31.2 & 30.9 & 53.6 \\
25\% shuffle & cell type & 1 & 7.03 & 56.5 & 33.1 & 32.0 & 56.4 \\
25\% shuffle & cell type & 2 & 9.04 & 54.9 & 32.6 & 29.5 & 49.3 \\
maximal shuffle & neuron & 0 & 1.92 & 33.9 & 18.1 & 17.5 & 32.1 \\
maximal shuffle & neuron & 1 & 2.37 & 35.1 & 18.5 & 16.3 & 29.9 \\
maximal shuffle & neuron & 2 & 1.84 & 16.3 & 9.3 & 8.6 & 14.9 \\
maximal shuffle & cell type & 0 & 3.85 & 44.1 & 23.7 & 21.9 & 40.6 \\
maximal shuffle & cell type & 1 & 4.66 & 27.6 & 16.0 & 12.0 & 22.1 \\
maximal shuffle & cell type & 2 & 0.98 & 15.0 & 7.9 & 7.5 & 13.9 \\
\bottomrule
\end{tabular}
\end{table}

\subsection*{S2.\enspace Results of the earlier study}
\label{sec:s-v1}
In an earlier paper \citep{chessfly_v1} we trained the same rate model on the male CNS compass wiring (neuPrint extraction, 434 neurons) and measured it against null networks, after 2,000 training updates. On the hold task, the median error of the real wiring was 1.7$^\circ$ (five seeds, 1.3--3.6$^\circ$), lower than 29/30 random rewirings and 24/30 degree-preserving rewirings; the network-level rank $p$ values were 0.065 and 0.23, so an advantage over degree-preserving wiring was not established. When angular velocity was delivered uniformly to all PEN neurons, the real wiring integrated (55.2$^\circ$ against 65.6$^\circ$ for ignoring turns), with all 30 degree-preserving rewirings again integrating better. 

Targeted rewiring of one-way connections disrupted holding (13.4$^\circ$ and 18.5$^\circ$ after moving 10\% and 20\% of all connections from the one-way pool), but moving the same number of reciprocal connections did not (2.0$^\circ$ and 1.5$^\circ$); the contrast reappeared in the hemibrain (21.3$^\circ$ against 1.3$^\circ$), but was explained by the effects of pathway strength changes. Hold error was also non-monotonic in the fraction of degree-preserving swaps: 25--60\% shuffles impaired holding more than maximal shuffles. Our present pilot hints at a reason: heavily shuffled wiring makes heading live on a deformed loop, an approximate attractor that preserves heading over short tests, rather than on the round ring of the real wiring (Sec.~\ref{sec:r3}, Table~\ref{tab:s-pilotmech}).  Intermediate shuffles break the ring without leaving an effective loop. The wiring-map distances for the rewired circuits of earlier experiments are used in Fig.~\ref{fig:2}B.

\subsection*{S3.\enspace Positions, neuron maps and distances}
\label{sec:s-maps}

\begin{table}[h]
\centering
\small
\caption{\textbf{Wiring-map distances by source of difference}, with the noise floor and the left--right control (random halves, five repeats; SD of the edge-removal control 0.003).}
\label{tab:distances}
\begin{tabular}{@{}lll r@{}}
\toprule
Source & Kind & Comparison & Distance \\
\midrule
Filtering & same fly & male CNS Codex vs.\ neuPrint & 0.022 \\
 & same fly & FlyWire filtered vs.\ unfiltered & 0.038 \\
Proofreading & same fly & male CNS v0.9 vs.\ v1.0 & 0.000 \\
 & same fly & BANC v626 vs.\ v888 & 0.104 \\
Synapse detector & same fly & FlyWire Princeton vs.\ Buhmann & 0.090 \\
 & same fly & FlyWire unfiltered vs.\ Buhmann & 0.084 \\
Between flies & same pipeline family & male CNS vs.\ hemibrain & 0.117 \\
 & different pipelines & hemibrain vs.\ FlyWire & 0.108 \\
 & different pipelines & FlyWire vs.\ BANC & 0.125 \\
 & different pipelines & male CNS vs.\ FlyWire & 0.137 \\
 & different pipelines & hemibrain vs.\ BANC & 0.185 \\
 & different pipelines & male CNS vs.\ BANC & 0.212 \\
\midrule
Noise floor & same fly & 10\% random edge removal (male CNS) & 0.032 \\
Left vs.\ right & same fly & male CNS (random halves 0.068) & 0.060 \\
 & same fly & BANC (random halves 0.111) & 0.100 \\
\bottomrule
\end{tabular}
\end{table}
\begin{table}[H]
\centering
\small
\caption{\textbf{Validation of compass positions inferred from wiring} against glomerulus labels (circular agreement; shuffled labels as control).}
\label{tab:s-positions}
\begin{tabular}{@{}lllrr@{}}
\toprule
Fly & Cell type & Method & Agreement & Shuffled \\
\midrule
male CNS & EPG & MDS of connectivity profiles & 0.99 & 0.26--0.31 \\
male CNS & PEN\_a & anchored to EPG & 1.00 & 0.36--0.49 \\
male CNS & PEN\_b & anchored to EPG & 1.00 & 0.33--0.40 \\
male CNS & PEG & anchored to EPG & 1.00 & 0.30--0.38 \\
hemibrain & EPG & MDS of connectivity profiles & 0.99 & 0.25--0.30 \\
hemibrain & PEN\_a & anchored to EPG & 1.00 & 0.31--0.35 \\
hemibrain & PEN\_b & anchored to EPG & 0.99 & 0.24--0.29 \\
hemibrain & PEG & anchored to EPG & 1.00 & 0.27--0.55 \\
\bottomrule
\end{tabular}
\end{table}

\begin{table}[H]
\centering
\small
\caption{\textbf{Exact glomerulus match of the neuron maps} between the two labelled flies (target $\rightarrow$ source).}
\label{tab:s-mapsval}
\begin{tabular}{@{}llr@{}}
\toprule
Map & Cell type & Exact match \\
\midrule
hemibrain $\rightarrow$ male CNS & EPG & 100\% \\
hemibrain $\rightarrow$ male CNS & PEN\_a & 100\% \\
hemibrain $\rightarrow$ male CNS & PEN\_b & 100\% \\
hemibrain $\rightarrow$ male CNS & PEG & 83\% \\
male CNS $\rightarrow$ hemibrain & EPG & 100\% \\
male CNS $\rightarrow$ hemibrain & PEN\_a & 100\% \\
male CNS $\rightarrow$ hemibrain & PEN\_b & 100\% \\
male CNS $\rightarrow$ hemibrain & PEG & 83\% \\
\bottomrule
\end{tabular}
\end{table}

\begin{table}[H]
\centering
\small
\caption{\textbf{Neuron maps for all ordered fly pairs}: correlation of log connectivity with used frame, with or without mirror frame, and fraction of source that is used.}
\label{tab:s-maps}
\begin{tabular}{@{}llrcr@{}}
\toprule
Source & Target & Alignment $r$ & Mirrored & Neurons used \\
\midrule
male CNS & hemibrain & 0.804 & no & 99\% \\
male CNS & FlyWire & 0.572 & no & 96\% \\
male CNS & BANC & 0.515 & no & 94\% \\
hemibrain & male CNS & 0.804 & no & 94\% \\
hemibrain & FlyWire & 0.570 & yes & 93\% \\
hemibrain & BANC & 0.528 & no & 93\% \\
FlyWire & male CNS & 0.581 & yes & 96\% \\
FlyWire & hemibrain & 0.563 & yes & 99\% \\
FlyWire & BANC & 0.434 & yes & 96\% \\
BANC & male CNS & 0.515 & yes & 91\% \\
BANC & hemibrain & 0.529 & no & 96\% \\
BANC & FlyWire & 0.423 & yes & 93\% \\
\bottomrule
\end{tabular}
\end{table}

\begin{table}[H]
\centering
\footnotesize
\caption{\textbf{Robustness of the decomposition to the distance measure.} Angular wiring-map distance (main text) and three standard measures: neuron-level input-fraction dissimilarity (Weights) and Jaccard distance between edge sets (both on direction-tuned neurons, matched by identity within a fly and by rank maps between flies), and cell-type pathway-share distance (Shares). Spearman correlation with the map distance across the 12 comparisons: Weights 0.93, Jaccard 0.77, Shares 0.93 (all $p \leq 0.004$).}
\label{tab:s-altmetrics}
\begin{tabular}{@{}lllrrrrr@{}}
\toprule
Source & Comparison & Matching & $n$ & Map & Weights & Jaccard & Shares \\
\midrule
filtering & male CNS vs male CNS neuPrint & neuron ID & 152 & 0.022 & 0.003 & 0.121 & 0.013 \\
filtering & FlyWire vs FlyWire unfilt. & neuron ID & 155 & 0.038 & 0.061 & 0.223 & 0.023 \\
proofreading & male CNS v0.9 vs male CNS & neuron ID & 152 & 0.000 & 0.000 & 0.000 & 0.000 \\
proofreading & BANC v626 vs BANC & neuron ID & 140 & 0.104 & 0.082 & 0.132 & 0.049 \\
synapse detector & FlyWire vs FlyWire Buhmann & neuron ID & 155 & 0.090 & 0.358 & 0.257 & 0.072 \\
synapse detector & FlyWire unfilt. vs FlyWire Buhmann & neuron ID & 155 & 0.084 & 0.304 & 0.336 & 0.065 \\
between flies & male CNS vs hemibrain & rank map & 152 & 0.117 & 0.542 & 0.293 & 0.085 \\
between flies & hemibrain vs FlyWire & rank map & 155 & 0.108 & 0.803 & 0.570 & 0.090 \\
between flies & FlyWire vs BANC & rank map & 151 & 0.125 & 1.107 & 0.646 & 0.097 \\
between flies & male CNS vs FlyWire & rank map & 155 & 0.137 & 0.826 & 0.561 & 0.075 \\
between flies & hemibrain vs BANC & rank map & 151 & 0.185 & 0.974 & 0.542 & 0.105 \\
between flies & male CNS vs BANC & rank map & 151 & 0.212 & 1.037 & 0.547 & 0.137 \\
\bottomrule
\end{tabular}
\end{table}

\subsection*{S4.\enspace Transplant matrices}
\label{sec:s-transplants}
Hold errors (degrees, 40-step test) for all pairs are in Tables~\ref{tab:s-s4}--\ref{tab:s-s4c}; rows are the wiring a network was trained on, columns the wiring it was tested on, the diagonal is the own-wiring error. Failed networks (seed 2 of male CNS and hemibrain in clean training) are excluded. For pairwise zero-shot and refit errors in Fig.~\ref{fig:5}, see \texttt{stage4d\_transplants.csv} in the repository.

\begin{table}[H]
\centering
\small
\caption{\textbf{Stage~4 transplants, cell-type gains} (clean training, 3--5 networks per row).}
\label{tab:s-s4}
\begin{tabular}{@{}lrrrr@{}}
\toprule
Trained on & male CNS & hemibrain & FlyWire & BANC \\
\midrule
male CNS & 1.2 & 89.4 & 90.7 & 89.2 \\
hemibrain & 87.9 & 0.9 & 86.2 & 90.7 \\
FlyWire & 91.1 & 91.7 & 0.7 & 88.6 \\
BANC & 89.3 & 88.5 & 83.9 & 1.1 \\
\bottomrule
\end{tabular}
\end{table}

\begin{table}[H]
\centering
\small
\caption{\textbf{Stage~4 transplants, per-neuron gains} (clean training, 3--5 networks per row).}
\label{tab:s-s4n}
\begin{tabular}{@{}lrrrr@{}}
\toprule
Trained on & male CNS & hemibrain & FlyWire & BANC \\
\midrule
male CNS & 0.7 & 82.0 & 91.1 & 88.0 \\
hemibrain & 82.2 & 0.6 & 88.2 & 90.2 \\
FlyWire & 91.0 & 91.2 & 0.5 & 89.6 \\
BANC & 87.3 & 89.1 & 90.1 & 0.5 \\
\bottomrule
\end{tabular}
\end{table}

\begin{table}[H]
\centering
\small
\caption{\textbf{Noise sweep}: Transplant matrices at each training-noise level (per-neuron gains, three seeds per row; two failed clean networks excluded).}
\label{tab:s-s4c}
\begin{tabular}{@{}llrrrr@{}}
\toprule
Level & Trained on & male CNS & hemibrain & FlyWire & BANC \\
\midrule
0 & male CNS & 0.7 & 78.7 & 89.0 & 88.1 \\
 & hemibrain & 77.6 & 0.7 & 89.4 & 90.3 \\
 & FlyWire & 90.6 & 89.4 & 0.5 & 90.0 \\
 & BANC & 87.4 & 87.1 & 87.3 & 0.4 \\
0.25 & male CNS & 7.0 & 78.5 & 87.8 & 88.1 \\
 & hemibrain & 57.0 & 17.9 & 88.7 & 87.5 \\
 & FlyWire & 91.0 & 91.1 & 7.6 & 88.6 \\
 & BANC & 91.6 & 91.0 & 62.1 & 8.1 \\
0.5 & male CNS & 8.7 & 60.4 & 83.5 & 88.2 \\
 & hemibrain & 47.1 & 15.8 & 89.3 & 88.1 \\
 & FlyWire & 91.0 & 91.2 & 14.8 & 87.7 \\
 & BANC & 91.3 & 90.5 & 84.0 & 10.5 \\
1 & male CNS & 21.0 & 69.0 & 83.7 & 86.9 \\
 & hemibrain & 57.4 & 24.4 & 83.9 & 87.2 \\
 & FlyWire & 90.9 & 91.3 & 19.9 & 86.4 \\
 & BANC & 91.6 & 91.6 & 83.1 & 20.5 \\
\bottomrule
\end{tabular}
\end{table}

\subsection*{S5.\enspace Statistics and mechanism measures}
\label{sec:s-stats}

\begin{table}[H]
\centering
\scriptsize
\caption{\textbf{All tests.} Prespecified tests (pre.) and exploratory tests (Expl.), with uncorrected $p$ and Holm-corrected $p$ within the prespecified family.}
\label{tab:s-stats}
\begin{tabular}{@{}p{3.0cm}p{6.0cm}p{3.4cm}rr@{}}
\toprule
Family & Test & Statistic & $p$ & $p_{\mathrm{Holm}}$ \\
\midrule
H1 transplant (pre.) & median transplant ratio R, celltype gains & R = 87 & -- & -- \\
H1 transplant (pre.) & median transplant ratio R, neuron gains & R = 162 & -- & -- \\
H2 level (pre.) & neuron-level R $>$ cell-type R (paired Wilcoxon, 12 pairs) & median 162 vs 87 & 2.4$\times10^{-4}$ & 7.3$\times10^{-4}$ \\
H3 distance (pre.) & $\rho$(wiring distance, R), celltype gains & $\rho$ = -0.33 & 0.85 & 0.85 \\
H3 distance (pre.) & $\rho$(wiring distance, R), neuron gains & $\rho$ = +0.25 & 0.21 & 0.42 \\
H4 mechanism (pre.) & banc celltype: networks meeting ring + fast return & 3/5 & -- & -- \\
H4 mechanism (pre.) & banc neuron: networks meeting ring + fast return & 5/5 & -- & -- \\
H4 mechanism (pre.) & fafb celltype: networks meeting ring + fast return & 4/5 & -- & -- \\
H4 mechanism (pre.) & fafb neuron: networks meeting ring + fast return & 5/5 & -- & -- \\
H4 mechanism (pre.) & hemibrain celltype: networks meeting ring + fast return & 3/4 & -- & -- \\
H4 mechanism (pre.) & hemibrain neuron: networks meeting ring + fast return & 4/4 & -- & -- \\
H4 mechanism (pre.) & mcns celltype: networks meeting ring + fast return & 3/4 & -- & -- \\
H4 mechanism (pre.) & mcns neuron: networks meeting ring + fast return & 4/4 & -- & -- \\
H4 mechanism (pre.) & positive control: Delta7 silencing breaks H4 & 26/31; error up in 100\% & -- & -- \\
H5 drift (descriptive) & fly E-PG drift, darkness & D = 0.0465 [0.0127, 0.1032] rad$^2$/s & -- & -- \\
H5 drift (descriptive) & fly E-PG drift, bright cue (jitter floor) & D = 0.0228 [0.0107, 0.0350] rad$^2$/s & -- & -- \\
H5 drift (descriptive) & fly E-PG drift, darkness minus bright & D = 0.0237 [-0.0103, 0.0763] rad$^2$/s & -- & -- \\
Expl.: fragility & pilot: error with all weights $\pm$1\% (median over networks) & 3.0 $\rightarrow$ 21.1 deg & -- & -- \\
Expl.: noise trade-off & $\rho$(noise level, own error) & $\rho$ = +0.83 & 9.6$\times10^{-13}$ & -- \\
Expl.: noise trade-off & $\rho$(noise level, error at weights x0.95) & $\rho$ = -0.86 & 9.9$\times10^{-15}$ & -- \\
Expl.: global gain & transplant error before vs after best global gain (median) & 87 $\rightarrow$ 83 deg & 6.3$\times10^{-5}$ & -- \\
Expl.: refit by source & 2 same fly: filtering: zero-shot -$>$ refit (median), rescued $<$30 deg & 45 $\rightarrow$ 8 deg, 92\% & -- & -- \\
Expl.: refit by source & 3 same fly: synapse detector: zero-shot -$>$ refit (median), rescued $<$30 deg & 83 $\rightarrow$ 61 deg, 33\% & -- & -- \\
Expl.: refit by source & 4 different flies: zero-shot -$>$ refit (median), rescued $<$30 deg & 88 $\rightarrow$ 57 deg, 21\% & -- & -- \\
Expl.: refit by source & refit error differs across categories (Kruskal-Wallis) & -- & 9.9$\times10^{-7}$ & -- \\
Expl.: refit by source & same-fly detector vs different flies, refit error (Mann-Whitney, two-sided) & 61 vs 57 deg & 0.2 & -- \\
Expl.: refit by source & $\rho$(wiring distance, refit error), 20 ordered pairs & $\rho$ = +0.67 & 0.001 & -- \\
\bottomrule
\end{tabular}
\end{table}

\begin{table}[H]
\centering
\footnotesize
\caption{\textbf{Mechanism measures of the Stage~4 networks} (trained networks only): H4 count, slowest Jacobian multiplier (median and range), radius variation and winding number of the trajectory in the top two principal components (exploratory), and hold error at 200 steps (degrees).}
\label{tab:s-h4}
\begin{tabular}{@{}llrrrrrrr@{}}
\toprule
Fly & Gains & $n$ & H4 & $\lambda_1$ med. & $\lambda_1$ range & Radius var. & $|$Winding$|=1$ & Err.\ 200 \\
\midrule
BANC & cell type & 5 & 3 & 1.000 & 0.998--1.009 & 0.21 & 5/5 & 12.0 \\
BANC & per neuron & 5 & 5 & 1.001 & 0.991--1.001 & 0.15 & 5/5 & 2.0 \\
FlyWire & cell type & 5 & 4 & 1.003 & 1.000--1.044 & 0.21 & 4/5 & 8.7 \\
FlyWire & per neuron & 5 & 5 & 1.001 & 1.000--1.003 & 0.21 & 5/5 & 3.5 \\
hemibrain & cell type & 4 & 3 & 1.014 & 1.008--1.021 & 0.30 & 4/4 & 12.7 \\
hemibrain & per neuron & 4 & 4 & 1.001 & 1.000--1.002 & 0.27 & 4/4 & 4.2 \\
male CNS & cell type & 4 & 3 & 0.998 & 0.996--1.000 & 0.11 & 4/4 & 20.8 \\
male CNS & per neuron & 4 & 4 & 1.002 & 1.000--1.002 & 0.15 & 4/4 & 4.8 \\
\bottomrule
\end{tabular}
\end{table}

\subsection*{S6.\enspace Preregistered hypotheses and outcomes}
\label{sec:s-prereg}
Table~\ref{tab:s-prereg}  provides an overview of preregistered hypotheses (doi:10.5281/zenodo.23382486) and their outcomes (deviations in Sec.~\ref{sec:m-prereg}).
\begin{table}[h]
\centering
\small
\caption{\textbf{Preregistered hypotheses and outcomes.}}
\label{tab:s-prereg}
\begin{tabular}{@{}p{3.2cm}p{5.2cm}p{6.3cm}@{}}
\toprule
Hypothesis & Prediction & Outcome \\
\midrule
H1 transfer & Robust ($R < 2$) or fragile ($R > 5$) transfer between flies & Fragile: median $R = 87$ (cell-type gains, registered) and 162 (per-neuron gains) \\
H2 parameter level & Neuron-level parameters transfer worse than cell-type parameters & Supported: $R$ larger at neuron level in 12 of 12 pairs, $p = 2.4\times10^{-4}$ (Holm $7.3\times10^{-4}$) \\
H3 distance & Transfer failure increases with wiring distance & Not supported ($\rho = -0.33$ and $+0.25$); all transplants at chance (floor effect) \\
H4 mechanism & Real wirings form a ring attractor with fast return (at least 3 of 5 seeds) & Supported in all 8 fly $\times$ parameterization combinations; 31 of 36 networks \\
H5 drift & Descriptive comparison of model drift in darkness with the DANDI 000289 estimate; no directional prediction & Fly drift reported (darkness $D = 0.047$, bright cue 0.023, excess 0.024 rad$^2$/s, $-0.010$ to 0.076); model comparison not made (deviation 8) \\
\bottomrule
\end{tabular}
\end{table}

\subsection*{S7.\enspace Validation of the drift method}
\label{sec:s-drift}
We tested the short-timescale fitting method on simulated phase bumps with known diffusion coefficients (0.002--0.1~rad$^2$/s) and Gaussian measurement noise (noise SD 0.1 or 0.3 rad). Replicate simulations consisted of 30 runs of 5-15 s sampled at 0.1 s. The set of runs was pooled and the $\mathrm{MSD}(t) = 2Dt + 2\sigma^2$ for lags up to 3 s, as for the fly data, with 20 replicates per condition. The fitted $D$ differed from the true value by $-5.2\%$ to $+2.2\%$  on average, and the fitted noise level by 0.088-0.303 rad for true values of 0.1 and 0.3 rad  (Table~\ref{tab:s-driftval}). The procedure is therefore unbiased over the range of fly estimates; the variation over replicates gives the precision that is achievable with a few minutes of data.

\begin{table}[H]
\centering
\small
\caption{\textbf{Test of the drift-method on simulated data } (20 replicates per condition).}
\label{tab:s-driftval}
\begin{tabular}{@{}lrrrrr@{}}
\toprule
True $D$ & True $\sigma$ & Fitted $D$ (mean) & SD & Bias & Fitted $\sigma$ \\
\midrule
0.002 & 0.1 & 0.0020 & 0.0002 & +1.2\% & 0.100 \\
0.002 & 0.3 & 0.0020 & 0.0007 & +0.6\% & 0.300 \\
0.01 & 0.1 & 0.0102 & 0.0012 & +2.2\% & 0.099 \\
0.01 & 0.3 & 0.0101 & 0.0021 & +1.2\% & 0.302 \\
0.04 & 0.1 & 0.0379 & 0.0060 & -5.2\% & 0.102 \\
0.04 & 0.3 & 0.0393 & 0.0070 & -1.7\% & 0.300 \\
0.1 & 0.1 & 0.0994 & 0.0176 & -0.6\% & 0.088 \\
0.1 & 0.3 & 0.0974 & 0.0135 & -2.6\% & 0.303 \\
\bottomrule
\end{tabular}
\end{table}

\end{document}